\documentclass[11pt]{article}
\usepackage{coling2020}
\usepackage{times}
\usepackage{url}
\usepackage{latexsym}
\usepackage{amsmath}
\usepackage{xcolor}
\usepackage{soul}
\usepackage{cleveref}
\usepackage{graphicx}
\usepackage{wrapfig,booktabs}
\usepackage{tabularx}
\usepackage{enumitem}
\usepackage{colortbl}
\usepackage{longtable}
\usepackage[font=small,labelfont=bf]{caption}
\newcolumntype{P}[1]{>{\centering\arraybackslash}p{#1}}
\crefname{section}{§}{§§}
\Crefname{section}{§}{§§}
\usepackage{floatrow}
\newfloatcommand{capbtabbox}{table}[][\FBwidth]

\usepackage[compact]{titlesec}
\colingfinalcopy % Uncomment this line for the final submission

\title{Adapting a Language Model for Controlled Affective Text Generation}
\author{Ishika Singh \thanks{\quad The two authors contributed equally to this work.}\qquad Ahsan Barkati\footnotemark[1] \qquad Tushar Goswamy \qquad \textbf{Ashutosh Modi} \\ 
Indian Institute of Technology Kanpur (IITK) \\
{\tt \{ishikas, ahsanb, tgoswamy\}@iitk.ac.in}\\ {\tt {ashutoshm}@cse.iitk.ac.in} 
\\}
  
\date{}

\begin{document}
\maketitle
\begin{abstract}
%Messages for human conversation are best conveyed by flavouring the sentences with emotional words. 
Human use language not just to convey information but also to express their inner feelings and mental states. In this work, we adapt the state-of-the-art language generation models to generate affective (emotional) text. We posit a model capable of generating affect-driven and topic focused sentences without losing grammatical correctness as the affect intensity increases. We propose to incorporate emotion as prior for the probabilistic state-of-the-art text generation model such as GPT-2. The model gives a user the flexibility to control the category and intensity of emotion as well as the topic of the generated text. Previous attempts at modelling fine-grained emotions fall out on grammatical correctness at extreme intensities, but our model is resilient to this and delivers robust results at all intensities. %We utilize the recently curated dataset called 
%NRC Emotion/Affect Intensity Lexicon \cite{LREC18-AIL} to incorporate affect category as well the intensity. 
We conduct automated evaluations and human studies to test the performance of our model, and provide a detailed comparison of the results with other models. In all evaluations, our model outperforms existing affective text generation models. 
% \iffalse
% \st{This is followed by demonstrating an application of the language model for story generation, advertisements and conversational agents for therapy chatbots.
% We aim to construct a emotion-controlled conversational agent which assesses the emotion of the user and it's intensity, and replies with a contrasting emotion at coarse and fine-grained level using the modelled text generator. }
% \fi
\end{abstract}

\section{Introduction}
\label{sec:intro}

\blfootnote{
    %
    % for review submission
    %
    % \hspace{-0.65cm}  % space normally used by the marker
    % Place licence statement here for the camera-ready version. See
    % Section~\ref{licence} of the instructions for preparing a
    % manuscript.
    %
    % % final paper: en-uk version 
    %
    % \hspace{-0.65cm}  % space normally used by the marker
    % This work is licensed under a Creative Commons 
    % Attribution 4.0 International Licence.
    % Licence details:
    % \url{http://creativecommons.org/licenses/by/4.0/}.
    % 
    % % final paper: en-us version 
    %
    \hspace{-0.65cm}  % space normally used by the marker
    This work is licensed under a Creative Commons 
    Attribution 4.0 International License.
    License details:
    \url{http://creativecommons.org/licenses/by/4.0/}.
}

 \noindent Affect (emotion) in language plays a critical role in conveying mental and emotional states, along with the  information intended to be conveyed. Machine Learning (ML) based text generation models are focused on minimising the error in the generated text by maintaining grammatical correctness, this often results in the generation of monotonous and dull conversations. Since current ML based models are trained on large corpora without any explicit affective information, they are unable to capture the emotional aspects of conversations  explicitly \cite{Asghar2018}. 
 %This happens because when traditional models are trained on a huge corpora,
 
% There needs to be automated processing and generation of affect-driven language, to improve the quality of conversations generated and make these models more empathetic towards a human user \cite{Colombo2019}.

There is a need for the automated processing and generation of affect-driven language. This will improve the quality of responses generated by a conversational system by making them  more empathetic towards a human user \cite{Colombo2019}. The role of controlled emotion in the generated text is to ensure more meaningful conversations between an AI-agent and a human. It is also aimed at establishing an emotional connect with the reader of the text. 
Such a model can be particularly useful for conversational therapy bots for generating appropriate emotional responses based on the user's mental state \cite{Spring:4525}. Affective generation model could also be useful in the development of interactive virtual agents \cite{janghorbani2019domain}. %\st{  Another application is story completion given the initial content, since emotions play a central role in one's experience of reading a novel and have consequences even after the book has been put down} \cite{Mar2011}. 
Affective advertisement generation is another area of application, especially for social advertisements aimed at inviting donations for a cause, where we need to emotionally appeal to the benefits of donation \cite{Moran2019}. Current research on incorporating language in robot learning \cite{luketina2019survey,bisk2020experience} can potentially benefit not just by enabling effective human-robot dialog, but also by optimizing other reinforcement learning based components. For instance, user's responses can be exploited to extract implicit emotion-based reward cues \cite{sumers2020learning}, and the robot can respond in contrasting emotion to gain user's trust. In interactive narrative generation \cite{ammanabrolu2020bringing}, an affect-incorporated language model will improve user's experience by making a long-lasting impact \cite{Mar2011}.  %\IS{[For concrete applications:] Current research on incorporating language in robot learning \cite{luketina2019survey,bisk2020experience} can potentially be benefit not just by enabling effective human-robot dialog, but also in optimizing other reinforcement learning components. For instance, user's responses can be exploited to extract implicit emotion-based reward cues \cite{sumers2020learning}, and the robot can respond in contrasting emotion to gain user's trust. In interactive narrative generation \cite{ammanabrolu2020bringing}, an affect-incorporated language model will improve user's experience by making a long-lasting impact \cite{Mar2011}.}

%Our work focuses on integrating the affective language generation approach with the SOTA language models. 
In this paper, we propose controlling the output of a neural language generation models with affective information. % (We understand that an emotion not just belongs to a concrete category, but there's also an associateddegree  to  which  it  is  expressed  or  perceived.   Hence,)
Intuitively, an emotion cannot be captured just by a discrete category, but there's also an associated degree to which it is expressed or perceived. Hence, we want to generate text based on a given emotion category ($e$) and its intensity ($\beta$). In addition to this, we also want the generated sentences to fall under a topic ($t$). We can either extract $e$ and $\beta$ from the context of conversation/text or can allow the user to choose these parameters. While generating affective text, we want to ensure that grammatical correctness is not compromised even at high emotion intensities. We propose an algorithm that generates grammatically correct text by sampling valid sentences along with optimizing for emotion label-intensity factors. %($P(w)$) along with optimizing for emotion label-intensity factors.

In particular, we propose coarse and fine-grained affective text generation model, built on top of GPT-2 \cite{radford2019language}. Our model provides degrees of freedom in terms of the choice of the base text generation model, the emotion category (ranging over 8 basic emotions), with fine-grained control over emotion intensity for each category, and the topic of the generated text. We provide detailed results of our model and a comparison with the existing models to establish the improvement brought in by our approach. We evaluate our model against the baselines on grammatical correctness, perceived emotion and intensity both using automated methods and human-annotations. We clearly see that the quality of text generated by out model both in terms of perceived emotion and grammatical correctness is considerably better than the existing system: AffectLM \cite{ghosh-etal-2017-affect}. As observed in experiments (section \ref{sec:grammCorrect}), in the case of AffectLM, with the increase in emotion intensity, the model compensates by generating more affective words at the cost of drop in grammatical correctness. However, our model tries to generate text as aligned to the given controls as possible while adhering to the grammatical constraints. %The perceived intensity is not as evident as models like AffectLM \cite{ghosh-etal-2017-affect} \AM{Previous line is not clear}, since the model tries to generate text as aligned to the given controls as possible within the grammatical constraints. 
To the best of our knowledge, this is the first affective text generation model that incorporates 8 emotion categories in the text generation output. The model is robust in terms of grammatical correctness at high emotion intensities, which makes it highly reliable for a number of applications. We release the model implementation and user studies at the following GitHub repository: \url{https://github.com/ishikasingh/Affective-text-gen}   %\footnote{\IS{The code is available at https://github.com/ishikasingh/Affective-text-gen}}

\section{Related Work}
\label{sec:length}
%\IS{ \newcite{deRosis2000} presented the need to revise NLG methods to incorporate emotions, personality traits and other non-strictly rational aspects of conversation by studying an example from a medical setting. They argued that all phases to text generation should be influenced by emotional factors.}

 \noindent Recently, several advancements in language generation have been made. The Conditional Transformer Language
Model For Controllable Generation (CTRL) \cite{keskarCTRL2019} provides a transformer language model that is conditioned on control codes, which allow the user to control the domain and topic of generated sentences, as well as define the intended task (like question-answering and machine translation). However, the CTRL model only allows to control the topic and does not provide the flexibility to control the emotion of the generated text. The Plug and Play Language Models (PPLM) \cite{Dathathri2020Plug} combines a pre-trained language model like GPT-2 with attribute classifiers that guide text generation. It enables the user to control the topic, sentiment (positive/negative) and the strength of the influence of these attributes (using the stepsize of a gradient descent equation) for the generated sentences. The PPLM model only allows the option of positive/negative sentiments in the output and does not deal with varied emotions. Moreover, PPLM model fails to generate grammatical text, when emotion intensity is increased. %like anger, fear, joy \IS{, etc.}. \IS{we didn't say that pplm also fails on high intensity since it uses convergence rate as intensity control, we say that for affectlm.. just for consistency} 
%PPLM is limited to discrete topic categories and broadly talks about only two sentiment categories \IS{previous and this sentence look repeated}, 
In contrast, in our model we have an extended list of eight basic emotions along with provision for controlling the intensity associated with each emotion. For controlling the intensity, we use human-annotated word list from NRC-EIL lexicon \cite{LREC18-AIL}. 
\begin{figure}
  \centering
  \includegraphics[scale=0.22]{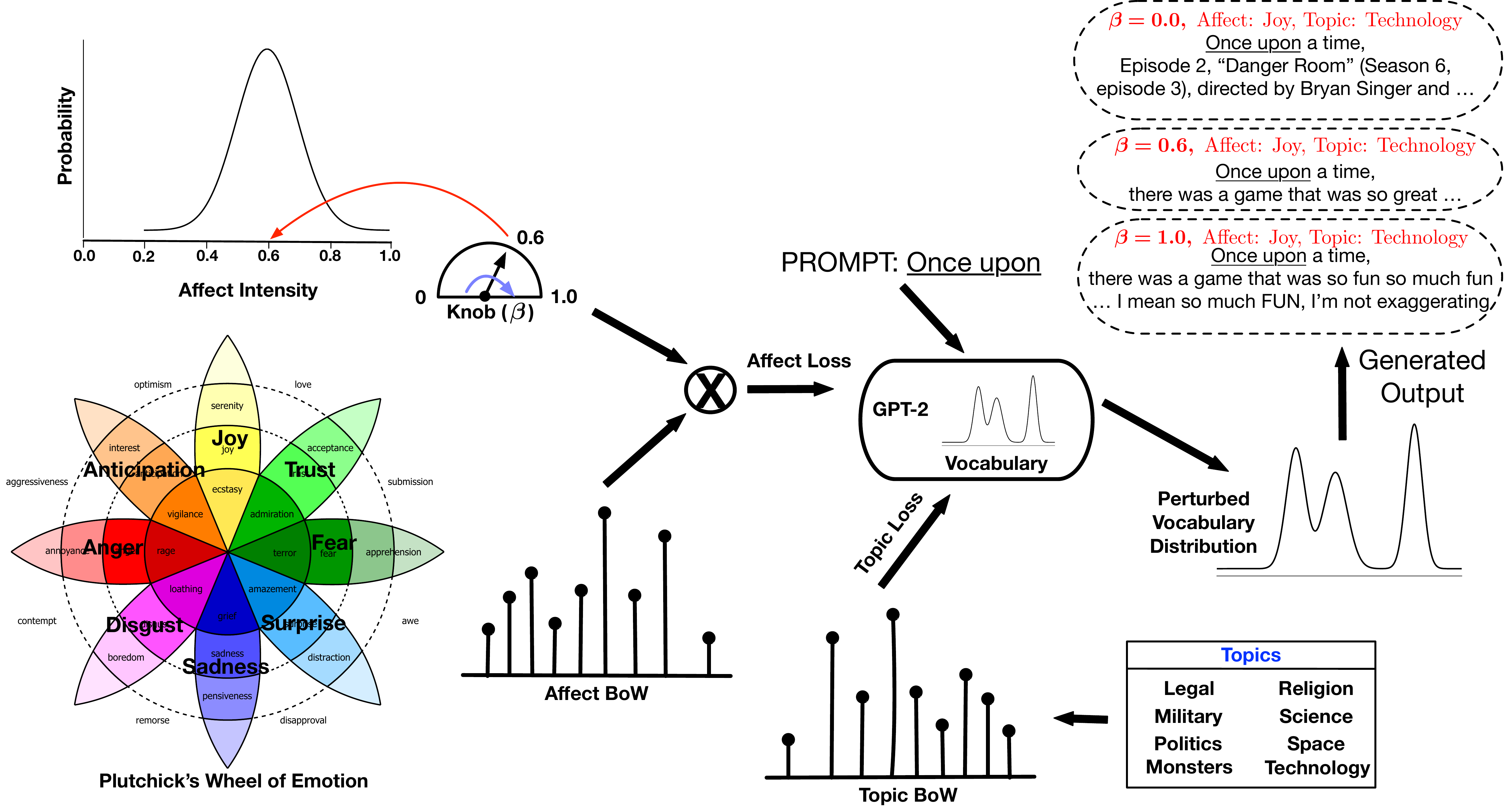}
  %\vspace{-0.3cm}
  \caption{\footnotesize{Architecture of the proposed model. Our model generates sentences when it receives a  Prompt, Topic, Emotion category and emotion intensity value from the user. Emotion Wheel image taken from wikipedia: \url{https://en.wikipedia.org/wiki/Emotion_classification}}}
  \label{fig:diagram}
   \vspace{-0.3cm}
\end{figure}
% \begin{figure}
%   \centering
%   \includegraphics[width=10cm, height=8cm]{./figs/Model_diagram.pdf}
%   \caption{Architecture of the proposed model. Our model generates sentences when it receives a  Prompt, Topic, Emotion category and emotion intensity value from the user}
%   \label{fig:diagram}
% \end{figure}

%One of the first attempt to incorporate emotions, personality traits and other non-strictly rational aspects of conversation into an NLG system was by \newcite{deRosis2000}. They studied an example from a medical setting and argued that all phases to text generation should be influenced by emotional factors.
In recent times, neural models for emotional text generation have been proposed. Affect-LM \cite{ghosh-etal-2017-affect} uses an LSTM-based approach for generating expressive emotional text. It is capable of generating sentences in 4 affect categories (Positive, Anxious, Sadness and Anger), and the affect intensity can be varied on a scale of 0 to $\infty$. However, since its introduction, several new text generating language models have been proposed (e.g., GPT-2 \cite{radford2019language}) which have outperformed previous RNN based language generation models.
%These State-of-the-Art (SOTA) text generative models include BERT \cite{DBLP:journals/corr/abs-1810-04805} and GPT-2 \cite{radford2019language}. 
The Affect-LM model depreciates in the grammatical correctness of its generated sentences as the affect intensity is increased to the higher end of the spectrum. Moreover, Affect-LM provides only 4 affect categories and misses out on emotions like surprise, anticipation, etc. In contrast, in our work we provide 8 basic emotion categories (Joy, Trust, Fear, Surprise, Sadness, Disgust, Anger and Anticipation). We base our choice of basic emotions on the theory proposed by Plutchik \cite{plutchik1962emotions,PLUTCHIK19803,Plutchik1994ThePA}, that argues that the eight basic emotions (largest proposed set) form four opposing pairs: joy–sadness, anger–fear, trust–disgust, and anticipation–surprise. In general, there are no clear boundaries proven between emotion categories, and it's perceived subjectively across individuals. 
Affective text generation models have been utilized in various applications.  \newcite{mahamood-reiter-2011-generating} present the application of affective NLG models to produce medical information summaries. These reports are communicated to the parents of babies in a Neonatal Intensive Care Unit, and need to have the appropriate affective tone so that they are able to deal with the emotional impact of the information. \newcite{mairesse-walker-2007-personage} present a system for language generation tailored on extroversion dimensions, which can be used for generating dialogues. An affect-driven dialogue system for generating emotional responses in a controlled manner has been designed by \newcite{Colombo2019}, that is aimed at making conversational systems more social. A significant number of research papers also deal with affective language models for chatbots in a commercial or therapy setting, such as the retrieval-based dialogue system by \newcite{f1764f31e341472d9f340f4e060786de}. \newcite{Chan2018} describe a sequence-to-sequence based emotional response generator for chatbots which are used in customer services, personal assistance and education. 

\setlength{\abovedisplayskip}{1pt}
\setlength{\belowdisplayskip}{1pt}

\section{Affective Model}
\subsection{Background}
Our model is based on the GPT-2 \cite{radford2019language} text generation model and the Plug and Play Language Model (PPLM) \cite{Dathathri2020Plug}. GPT-2 is a transformer-based language model which has shown superior performance on several NLP tasks, including text generation. The GPT-2 model generates text that is word by word conditioned on past context represented in the form of a history embedding matrix $H_{t}$. The model updates the history matrix recursively and samples the next word as,
\begin{align}
    O_{t+1}, H_{t+1}&=\mathrm{LM}\left(s_{t}, H_{t}\right)\\
    s_{t+1} \sim p_{t+1}&=\operatorname{Softmax}\left(W \cdot O_{t+1}\right)
\end{align}
where $O_{t+1}$ is the token embedding used to get the next word's probability distribution $p_{t+1}$ by learning a parameter $W$, and $s_{t+1}$ represents the sampled word at $(t+1)$th iteration. 

In order to efficiently update the model, as well as to generate controlled text we bring in the idea of alternating optimization or a projected gradient descent in the direction which optimizes the attribute category probability distribution $p_{t+1}(a|w)$ projected on sentence probability distribution $p_{t+1}(w)$ as done in PPLM. Here, $p_{t+1}(a|w)$ is the probability that $(t+1)$th word belongs to affect category $a$, and  $p_{t+1}(w)$ is the probability that $(t+1)$th word is a grammatically correct occurrence. PPLM allows to plug in any text generation model to optimize loss associated with $p_{t+1}(w)$. We use GPT-2 Medium since it has performed well for domain-specific text generation without the need to use domain-specific training datasets. PPLM then perturbs this $p_{t+1}(w$) (by changing the history matrix $H_t$) given by the plugged-in model, such that the generated text has a higher probability of belonging to a given attribute or topic, where the topics are represented by Bag of Words (BoW). The following Gradient Descent (GD) is performed to execute this perturbation,
\begin{align}
   H^{'}_{t} = H_t - \eta \frac{\partial  \text{Loss}}{\partial H_t} \label{GD}
\end{align}
such that we get the perturbed next word distribution $p^{'}_{t+1}$ as,
\begin{align}
    O^{'}_{t+1}, H_{t+1}&=\mathrm{LM}\left(s_{t}, H^{'}_{t}\right) \label{perturb}\\
    s_{t+1} \sim p^{'}_{t+1}&=\operatorname{Softmax}\left(W \cdot O^{'}_{t+1}\right) \label{sample}
\end{align}
  The Loss in Equation \ref{GD} has two terms: a KL-Divergence term, which keeps the perturbed next word probability distribution close to the actual one, hence ensuring grammatical correctness; and a loss associated with overall probability of words belonging to the given attribute. 
  \begin{align}
&\text{Loss} =\text{KLD}_{perturbed - unperturbed} + \text{Loss}_{topic}\\
&\text{Loss}_{topic} = -\log(\sum(BoW probs))
\end{align}

  In Equation 7, the $\sum{BoW probs}$ at time step $t$ is defined as $\sum_{i} p_t \cdot h_i$, where $h_i$ is a one-hot encoding of the $i^{th}$ word from the topic's bag of words. These two loss components, when used for two sequential steps of GD, relate to an alternating optimization technique. Combining these two losses to perform a single step of GD has a similar effect, hence it is used to minimize the loss by perturbing $H_t$. %The GD algorithm perturbs $H_t$ so as to minimize the defined loss. %this defined loss, formally written as,

\subsection{Proposed Model}
% \begin{figure}
%   \centering
%   \includegraphics[width=10cm, height=8cm]{./figs/Model_diagram.pdf}
%   \caption{Architecture of the proposed model. Our model generates sentences when it receives a  Prompt, Topic, Emotion category and emotion intensity value from the user}
%   \label{fig:diagram}
% \end{figure}

 The architecture of our model is shown in Figure \ref{fig:diagram}. The idea behind our model is similar to the above attribute perturbation in the sentences. We define a new loss term for the perturbation which steers the generation towards the required affect. It also provides an option to control the intensity of the emotion in the generated sentences. The new loss function is defined as:
\begin{align}
&\text{Loss} = \text{KLD Loss} + \text{Loss}_{topic}+ \text{Loss}_{affect}\label{5}\\
&\text{Loss}_{affect} = -\log((BoW probs)\cdot(\mathcal{N}(\text{affectInt}, knob, var))) \label{6}
\end{align}
% $\text{Loss}_{affect} = -\log((BOW probs)\cdot(\mathcal{N}(\text{affectInt}, knob, var)))$
In Equation \ref{6}, the $\mathcal{N}(\text{affectInt}, knob, var)$ is a Gaussian function, which is used to control the intensity of the affective text generation. Here 'affectInt' represents the intensity values for the BoWs corresponding to the emotion category, ranging from $0$ to $1$, where $1$ is the maximum intensity. The $knob$ (the mean of the Gaussian) scales-up the values for the words closer to it, and scales it down for those far away from the mean, hence increasing the probability of the words with intensity values closer to the $knob$ value. The $var$ provides with flexibility on the intensity range to be picked. The dot product of the scaled intensity scores with the BoW probabilities is to be maximized during the optimization. 
%It can be re-framed as a minimization objective by taking negative log of this dot product.
This new loss in Equation \ref{5} is then used to perturb the model history $H_t$, to increase the probability for words at $(t+1)$th iteration corresponding to the given emotion category and the intensity.

To incorporate affect, we use human-annotated affect information provided by NRC, which fulfils both BoWs - emotion categories and emotion intensities. 
% The NRC Emotion Lexicon \cite{Mohammad13}, \cite{mohammad-turney-2010-emotions} is a list of manually annotated English words and their associations with eight basic emotions (anger, fear, anticipation, trust, surprise, sadness, joy, and disgust) for 14,182 unigrams. 
The NRC Emotion Intensity Lexicon \cite{LREC18-AIL} provides real-valued intensity scores for approximately 10,000 words belonging to eight basic emotions (anger, fear, sadness, joy, anticipation, trust, surprise, trust).  The intensity values range from 0 to 1, 1 being highly belonging to the category and 0 being neutral of the category. This scale defines the range for our $knob$ in the loss defined in Equation \ref{6}.
% These scores are evaluated using Best-Worst Scaling, i.e., annotators were asked to label the best and worst belongingness of among a tuple of 4 words to a particular category (emotion here). These annotations were converted into ranks, which in turn were converted to scores between 0 to 1, 

% model hyperparam values 
We perturb the history as in Equation \ref{GD}, and feed it to the LM as in Equation \ref{perturb}. During the implementation, we perform 3 iterations of these two steps to accumulate enough perturbation in $H_t$ such that the loss is desirably reduced, and this is finally used to sample the next word as shown in Equation \ref{sample}. The convergence rate was set to $\eta = 0.005$.

%\subsection{Model as Affective Text Generator}
Our model can be used to generate affective sentences by choosing the desired emotion category. We support a varied range of 8 basic human emotions: joy, anger, fear, sadness, surprise, anticipation, disgust, and trust. Moreover, the user can select emotion intensity on a continuous scale of 0 to 1; where 0 stands for a neutral sentence and 1 for the highest possible intensity sentence (with grammatical correctness remaining intact) within the given category. The affect words are more likely to get sampled, which are closer to the given $knob$ value. 
% Figure \ref{fig:emocat} shows a snapshop of affect category words along with their distribution on intensity scale.
Some sentences generated by our model can be seen in Table \ref{examples}. These examples illustrate that our model is able to incorporate different emotion categories and retain the grammatical correctness in all sentences. We can also observe that the intensity conveyed by the sentences for the chosen emotion increases as we increase the knob value.

\begin{table*}[t]

\begin{center}
\begin{small}
\begin{tabular}{P{0.1\textwidth}|P{0.09\textwidth}|P{0.06\textwidth}|P{0.04\textwidth}|p{0.55\textwidth}}%{|c|c|c|c|l|}
\hline
\multicolumn{1}{c|}{\textbf{Prompt}} & \multicolumn{1}{c|}{\textbf{Topic}} & \multicolumn{1}{c|}{\textbf{Emotion}} & \multicolumn{1}{c|}{\textbf{Knob ($\beta$})} & \multicolumn{1}{c}{\textbf{Text}}\\
\hline
\multicolumn{5}{c}{\textbf{Our Model}}\\
\hline
%There exists & Science & Fear & 0.0 & There exists a vast, vast universe of material that is known as the material record, or MDR, that we can study and understand. \\

%There exists & Science & Fear & 0.6 & There exists a very good argument that the US has been using chemical weapons against the population since the 1970s. In the case of chemical weapons, the evidence is quite overwhelming. \\

%There exists & Science & Fear & 1.0 & There exists a very good argument that the world is going to end, or is about to end. It's a frightening idea, and I've always felt that we are all doomed. \\

%Once upon & Religion & Joy & 0.0 & Once upon a time, there were two types of people in America: those who were born into wealth, and those who were born into poverty. \\

%To conclude & Politics & Anger & 0.0 & To conclude, the evidence is clear that the US has the ability to make its own decisions regarding what to do with the oil and gas in the ground, which is why we do this.  \\

The President & Politics & Anger & 0.6 & The President's first budget is a huge tax cut for the rich and an increase in debt for our military. He's a disaster. — Senator Bernie Sanders (@SenSanders) December 17, 2017.. \\

The President & Politics & Anger & 1.0 & The President's first budget is a huge tax cut for the rich and an increase in debt for our military. He's a war criminal who wants to destroy our economy and take away our freedom.  \\

Once upon & Technology & Joy & 0.6 & Once upon a time, there was a game that was so great, so much fun you could play it over and over again.  \\

Once upon & Technology & Joy & 1.0 & Once upon a time, there was a game that was so fun, so much fun... I mean, so much FUN. I'm not exaggerating when I say that it was the best game in the world! \\
%I feel so & No Topic & Sadness & 0.0 & I feel so sad for the girls that are still in high school. I'm sure that they will be able to go on to college without any problem. But I don't want them to feel like they're doing nothing.  \\

I feel so & No Topic & Sadness & 0.6 & I feel so bad for the kid that was killed. The guy was a great guy. He was the kind of guy that you'd want to hang out with.  \\

I feel so & No Topic & Sadness & 1.0 & I feel so sad for the victims of the shooting. The news has been terrible, but we have to deal with it. I'm sorry that the shooting happened.  \\
\hline
\multicolumn{5}{c}{\textbf{AffectLM}}\\
\hline

The President & NA & Anger & 3 & the president of the country  you know . i i i don't know i just hope that it 's just . it 's just going to kill innocent people . i i i feel like it 's ridiculous for the rest of the united states especially in the united states
 \\

The President & NA & Anger & 5 & the president of the country  the fight against the united states hate the us and mass destruction attack and fight against killing you hate hate hate hate hate
  \\
Once upon & NA & Positive & 3 & once upon a year . i don't . oh . i think i i think i think ah i 'm not sure it 's like it 's the same way i 've seen it on the news . \\

Once upon & NA & Positive & 5 & once upon a time i just . i just do a lot of things with my family but it 's yeah it is my best friend is a . that 's not a good friend . yeah . yeah it it wouldn't be a good thing to do . \\
I feel so & NA & Sadness & 3 & i feel so  sad that you know emotional feelings are losing you know . oh yeah . as a result of . you know . of course . you know . you know \\

I feel so & NA & Sadness & 5 & i feel so  sad because i 've lost feelings lost feelings about losing feelings and feelings and feelings lose feelings feelings and feelings  \\

\hline
\end{tabular}
\end{small}
\end{center}
\caption{Examples of sentences generated by our model and AffectLM for different emotion categories, intensities and topics. More examples in \cref{supp:generation} and \cref{supp:affectlm}.} %\IS{should we give examples with same prompt-emotion pair?}} %\IS{More egs on emotion changing with const. prompt, intensity, etc.}}
\label{examples}
\vspace{-0.4cm}
\end{table*}

% \begin{figure}
%   \includegraphics[width=\linewidth]{./figs/emotion-word-dist (1).pdf}
%   \vspace{-1cm}
%   \caption{Emotion Category Word Snapshot with Intensity Scores}
%   \label{fig:emocat}
% \end{figure}

%\section{Corpus Description}

% \iffalse
% Since we are working on Statistical NLP, the models are learned from data. Corpus is an important component of the project. Key points (Following can be in any order):\\
% \begin{enumerate}
% \item Describe corpus/dataset that is going to be used, if it is mutlimodal describe each modality
% \item In case you are going to create data, how is data created or acquired? 
% \item Describe the key statistics of the data, train/val/test split or number of words or mean ,std, etc. whatever is applicable and will have an impact on the solution
% \item If you plan to augment this data with external data sources, mention that
% \end{enumerate}
% \fi

%%%%%%% Experiment and Result Evaluation %%%%%%%%

% \section{Model Training}
% \AM{Briefly write about how you trained the model, hyper-parameters, epochs, etc., write about examples of texts generated e.g. table 1, }

\section{Model Evaluation and Experiments} 
 \noindent We evaluated our model using complementary evaluation approaches: Automated evaluation and Human evaluation. Several experiments were conducted based on these evaluations, as described next.

\subsection{Automated Evaluations}
We quantitatively evaluate our model for perceived emotion intensity  and grammatical correctness of the generated text. To evaluate the emotional intensity of a sentence, we used a pre-trained emotion prediction model, the Affective Norms model\footnote{Affective Norms model (IMS EmoInt) was the second best performing model in the IMS Emotional Intensity Prediction task organized at WASSA workshop 2017} \cite{Koeper2017}. Given a sentence and its emotion (anger/fear/sadness/joy), the model evaluates the intensity felt by the speaker on a scale of 0 to 1. It is essentially a Random Forrest Regressor, trained on manually labelled tweets, which we have adapted for our generated sentences. To evaluate the grammaticality of the generated text, we used perplexity. We compared the text generated by the models against GPT \cite{Radford2018ImprovingLU} as the ground truth, this approach is similar to the one used for evaluating PPLM \cite{Dathathri2020Plug}.

\subsection{Human Evaluations}
We created $3$ different tasks for human evaluations comprising over 400 sentences. Two graduate students fluent in English (and oblivious to this project) performed these tasks and evaluated our model and other competitor models. %We collected annotations from 2 M.S. Computer Science students for all the following tasks. 
Task 1 and 2 were about sentence emotion classification with $4$ (positive emotion, anger, sadness, neutral) and $8$ (anger, fear, sadness, joy, anticipation, trust, surprise, trust) emotion classes respectively. Task 3 was aimed to study the grammatical correctness and perceived intensity.% trends for a given emotion intensity ($knob$) as input to the model. 

\noindent\textbf{Task 1} The annotators were presented with $72$ text snippets (consisting of at most 2 sentences) equally distributed over each of the $3$ classes (Sad, Angry, Positive Emotion) and generated by two models (AffectLM and our model). % and over each of the $3$ classes (Sad, Angry, Positive Emotion) of both models. 
Association between a text snippet and its generator (model) were unknown to the annotators. The sentences uniformly belonged $3$ intensity values ($\beta = 1, 2, 3$ for AffectLM and $knob = 0.4, 0.6, 1$ for our model). We gave `positive emotion' as one of the options, for a fair comparison between both the models since it is covered by both models\footnote{the `joy' category of our model is a subset of positive emotion class}. We also gave a `neutral' option, in case annotators felt that there is no emotion intended by the text, although all sentences were conditioned on one emotion category. The prediction accuracy was averaged over both the annotators.% calculated for both annotators and an average of the two numbers were noted. 

\noindent\textbf{Task 2} The annotators were given $96$ text snippets generated by our model, equally distributed over each of the $8$ emotion classes and $3$ intensity values ($knob = 0.4, 0.6, 1$). The prediction accuracy was averaged over both the annotators. %The accruracies were calculated for both annotators and an average of the two numbers were noted. \\

\noindent\textbf{Task 3} The annotators were given 39 sets of sentences. Each set comprised of 6 sentences, and the annotators were informed about the input emotion category of each sentence. %from any one of the emotion/sentiment categories, which the annotators were informed about. 
The annotators were asked to rate grammatical correctness for each sentence on a 7 point Likert scale. Moreover, they were asked to rank these sentence from 1 to 6 based on the relative intensity of the emotion expressed, 6 being the highest intensity rank. We curated 3 sets for each of the 3 AffectLM emotion categories (Sad, Angry, Positive Emotion), 2 sentiment categories (Positive, Negative) from PPLM, and 8 emotion categories from our model. The ratings from both the  annotators were averaged, each for grammatical correctness ratings and intensity ranks. %\AM{previous line is not clear}
% \iffalse
% To detect the emotion class of a sentence, we propose 2 approaches:
% \begin{enumerate}
%     \item IMS EmoInt: We can use the same model by getting the intensity scores under all 4 emotions for a generated sentence, and then assigning the emotion label which obtained the highest intensity score. 
%     \item NRC Lexicon: Another approach would be to utilise the NRC lexicon, which gives us a vocabulary of words in each emotion category with their intensities. We can compute the count of words belonging to each emotion, multiply the count with the intensities, and this score will be used to assign an emotion label to the sentence.
% \end{enumerate}
% \fi
%\\

We tested our model's performance in terms of its ability to generate text for a given emotion, the effectiveness towards adapting to emotional intensity and the impact of changing these two parameters on the grammatical correctness of the generated text. These experiments are described in the next section.

\iffalse
 \footnote{Model demo link:  \url{https://colab.research.google.com/drive/1mNadIY9gxyW6H4Ky3zPjxJKrLGLSKRtc}}

\begin{figure}[H]
    \centering
    \includegraphics[width=5cm, height=4cm]{project-MidSem-template/nlp1.PNG}
    \caption{Generated text for emotion: 'fear'}
    \label{fig:fear}
\end{figure}

\begin{figure}[H]
    \centering
    \includegraphics[width=5cm, height=4cm]{project-MidSem-template/nlp2.PNG}
    \caption{Generated text for emotion: 'joy'}
    \label{fig:fear}
\end{figure}
\fi

% \textbf{Perplexity score for (MOSI, MOSEI, MELD, SEMAINE) emotional text corpus} \textcolor{blue}{our vs. baseline}
% \\ \textcolor{blue}{table : datasets , model scores}\\ \\

\begin{figure}[htbp]
  \centering
  \includegraphics[width=\textwidth]{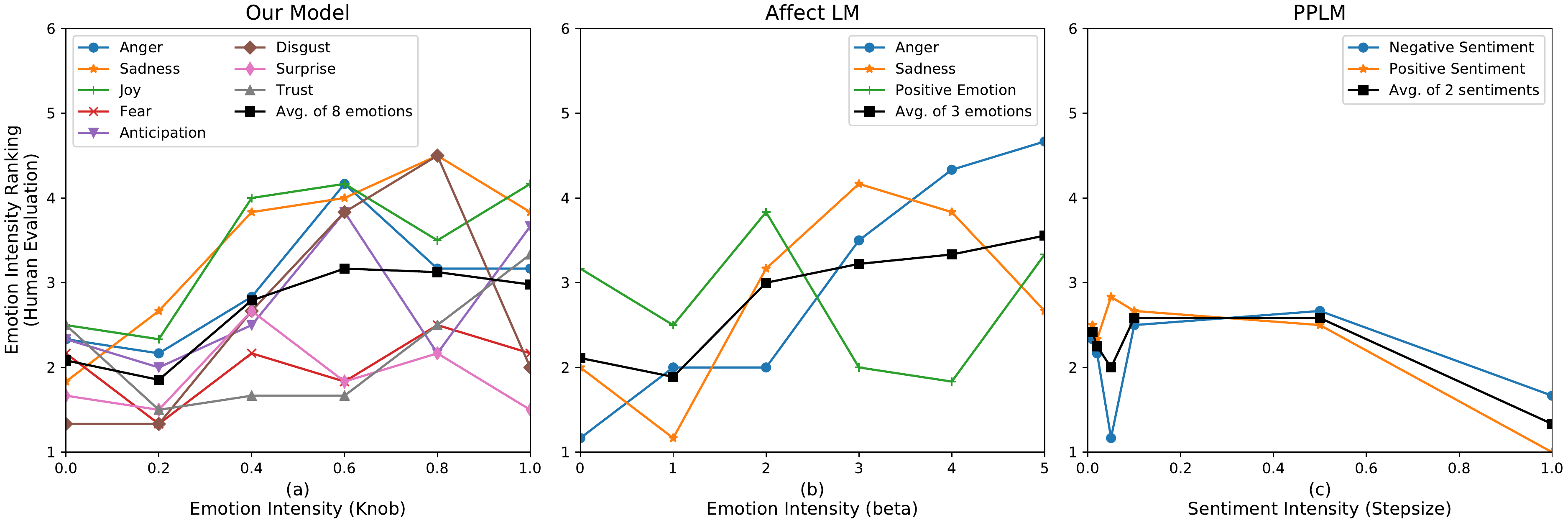}
  \caption{Human Perceived Intensity Evaluations}
  \label{hum-int}
  \vspace{-0.2cm}
\end{figure}

\subsection{Comparison Experiments} \label{sec:exp}
We compared our model's performance with AffectLM as the baseline model,  on 3 common emotion classes: Sad, Angry, Joy (our model)/Positive Emotion (AffectLM).

\noindent\textbf{True emotion vs predicted emotion:} This experiment was conducted via human evaluations. Annotations from Task 1 were used for this experiment. Table \ref{classif} shows the results.
%This experiment was conducted via human evaluations. Despite a lot of work done in automated emotion classification, best models perform as good as $70\%$ on classification accuracy (although much better on sentiment classification). We didn't want the error from pre-trained  models to reflect in our evaluations, hence we show results from human annotated emotion classes only for this experiment. Annotations from Task 1 was used for this experiment. Table \ref{classif} shows the results. 
% \vspace{-0.8cm}

\begin{figure}[htbp]
  \centering
  \includegraphics[width=\textwidth]{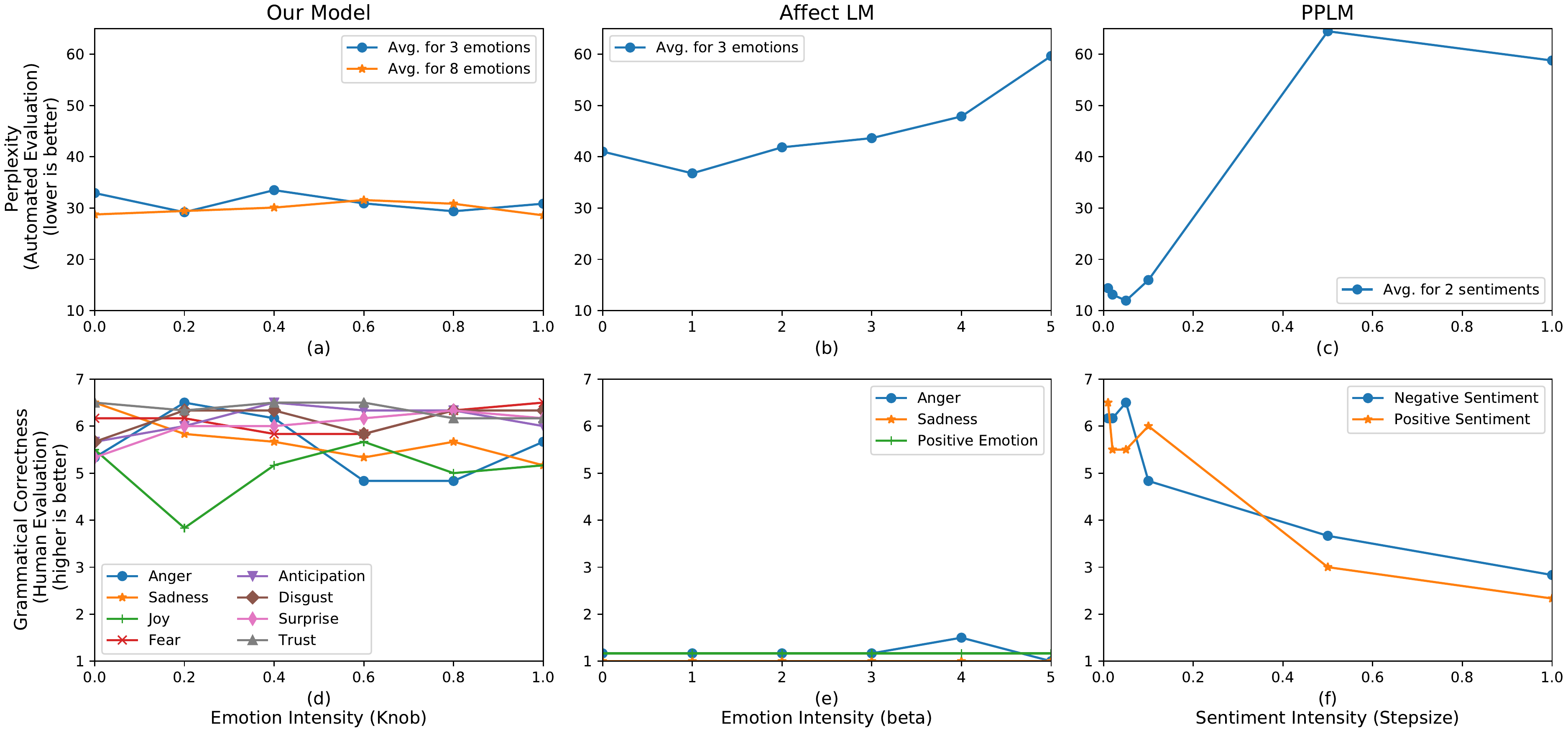}
  \caption{Grammatical Correctness Evaluations}
  \label{gram}
%   \vspace{-0.4cm}
\end{figure}

\noindent\textbf{Intensity and grammatical correctness trend for generated text with varying knob value:} 
\begin{itemize}[noitemsep,topsep=0pt]

\item Automated Evaluations:
Figure \ref{gram}(a), (b) shows the results for impact on perplexity (measure for grammatical correctness). Figure \ref{auto-int} shows perceived intensity with changing model intensity.
\item Human Evaluations: Annotations from Task 3 for three common emotion classes of AffectLM and our model were used for this experiment. Figure \ref{hum-int}(a), (b) shows the comparison for perceived intensity and \ref{gram}(d), (e) shows the comparison for grammatical correctness.
\end{itemize}

We compared our model with PPLM \cite{Dathathri2020Plug} on text quality with increasing intensity across our 8 emotion classes, and 2 sentiment classes (present in PPLM).

\noindent\textbf{Intensity and grammatical correctness trend for generated text with varying knob value:} 
\begin{itemize}[noitemsep,topsep=0pt]
 \item Automated Evaluations: Figure \ref{gram}(a), (c) shows the results for impact on perplexity (measure for grammatical correctness) with changing model intensity. 
    %As we can see, PPLM sees a sudden increase in perplexity value with actual intensity because of using the convergence parameter as intensity knob, while our model's perplexity is more or less a constant value with change in intensity.
    \item Human Evaluations: Annotations from Task 3 for 8 emotion classes of our model and 2 sentiment classes from PPLM were used for this experiment. Figure \ref{hum-int}(a), (c) shows the comparison for perceived intensity and \ref{gram}(d), (f) shows that for grammatical correctness.
\end{itemize}

\subsection{Independent Experiments}
We evaluated our model independently for 8 fine-grained emotion classes. 

\noindent\textbf{True emotion class vs predicted emotion:}
This experiment was conducted via human evaluations. Annotations from Task 2 were used for this experiment. Table \ref{classif} shows the results.

\noindent\textbf{Intensity and grammatical correctness trend for generated text with varying knob value:} 
\begin{itemize}[noitemsep,topsep=0pt]
    \item Automated Evaluations: Figure \ref{gram}(a) shows the results for impact on perplexity (measure for grammatical correctness) with changing model intensity. 
    \item Human Evaluations: Annotations from Task 3 for 8 emotion classes of our model were used for this experiment. Figure \ref{hum-int}(a) show the results for perceived intensity and \ref{gram}(d) show that for grammatical correctness.
\end{itemize}

  \begin{minipage}{\textwidth}
   \vspace{0.5cm}
  \begin{minipage}[b]{0.55\textwidth}
    \centering
    \includegraphics[width=\textwidth, height=4.5cm]{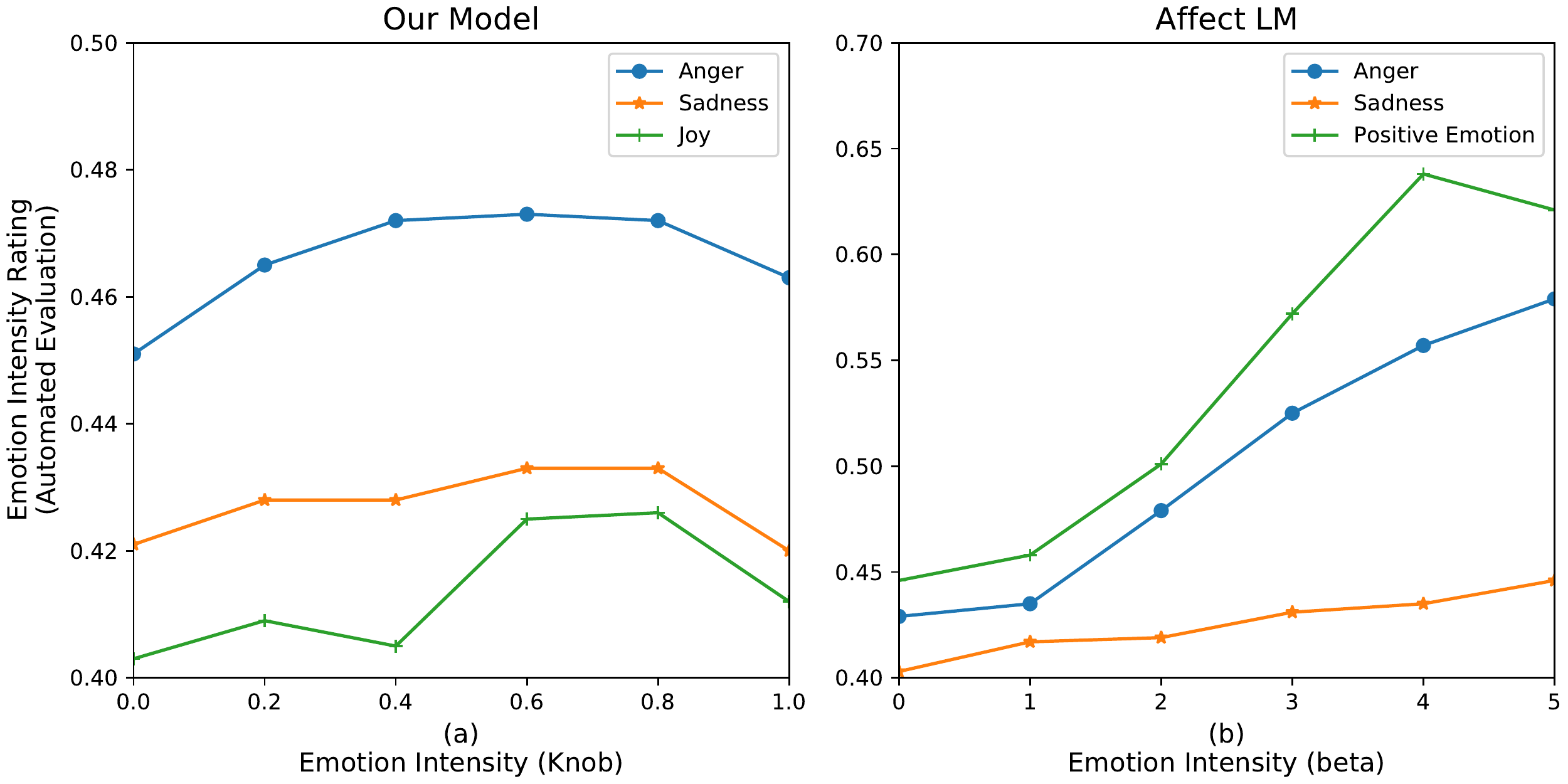}
    \captionof{figure}{Automated Perceived Intensity Evaluations%
  \label{auto-int}}
  \end{minipage}
  \hfill
  \begin{minipage}[b]{0.45\textwidth}
    \centering
   \begin{small}
  \begin{tabular}{|c|cc|}
\hline
$\beta$ (Knob) & \multicolumn{2}{|c|}{Emotion Prediction Accuracy} \\
\hline
\multicolumn{3}{|c|}{Affect LM}\\ \hline
   &\multicolumn{2}{c|}{(3 emotions)} \\ \hline
1   &\multicolumn{2}{c|}{0.392} \\
2   & \multicolumn{2}{c|}{0.542}\\
3   & \multicolumn{2}{c|}{0.542}\\
\hline
\multicolumn{3}{|c|}{Our Model}\\ \hline
% Knob & EPA & EPA\\
& (3 emotions) & (8 emotions)\\
\hline
0.4   & 0.833 & 0.359 \\
0.6   & 0.625 & 0.344\\
1.0   & 0.625 & 0.281\\
\hline
% \multicolumn{2}{|c|}{Our Model - 8 emotions}\\ \hline
% Knob & EPA \\
% \hline
% 0.4   & 0.359 \\
% 0.6   & 0.344\\
% 1.0   & 0.281\\
% \hline
\end{tabular}
\end{small}
      \captionof{table}{Human Evaluations on Classification Tasks (Task 1 and Task 2) \label{classif}}
    \end{minipage}
  \end{minipage}
%%%%%%%%%%%%%%%%%%%%%%%%%%%%%%%%%%%%% end new code %%%%%

%0.317, 0.010, 0.048, |0.002, 0.000, 0.0167, 0.352, 0.415,| 0.639, 0.203, 0.000, 0.188, 0.067
% %0.451, 0.146, 0.330, 0.553, -, 1, 0.135, 0.008, 0.749, 0.758, 0.692, 0.625, 0.671
% \begin{wraptable}{r}{0.45\textwidth}
% % \vskip -0.35in
% \begin{center}
% \begin{small}
% \begin{sc}
% \begin{tabular}{|c|c|}
% \hline
% \multicolumn{2}{|c|}{Affect LM for 3 emotions}\\ \hline
% $\beta$ & Emotion Prediction Accuracy \\
% \hline
% 1   & 0.392 \\
% 2   & 0.542\\
% 3   & 0.542\\
% \hline
% \multicolumn{2}{|c|}{Our Model for 3 emotions}\\ \hline
% Knob & Emotion Prediction Accuracy \\
% \hline
% 0.4   & 0.833 \\
% 0.6   & 0.625\\
% 1.0   & 0.625\\
% \hline
% \multicolumn{2}{|c|}{Our Model for 8 emotions}\\ \hline
% Knob & Emotion Prediction Accuracy \\
% \hline
% 0.4   & 0.359 \\
% 0.6   & 0.344\\
% 1.0   & 0.281\\
% \hline
% \end{tabular}
% \end{sc}
% \end{small}
% \end{center}
% \caption{Human Evaluations on Classification Tasks (Task 1 and Task 2)}
% \label{classif}
% \vspace{-1cm}
% \end{wraptable} 
\section{Analysis}
\noindent We conducted 26 ANOVA analysis with human-annotated predicted emotion intensity and grammatical correctness ratings as Dependent Variables (DVs), and knob value as Independent Variable (IV). The inter-agreement score was calculated among the two annotators for each of the tasks using Krippendorff's $\alpha$. For Task 1 and Task 2, we observed moderate agreement ($\alpha =0.54 $ and $\alpha =0.42 $ respectively), given the subjective nature of emotion perception. For Task 3, we observed great agreement ($\alpha =0.74 $) on grammatical correctness ratings, while we got $\alpha =0.23 $ on intensity rankings, showing the diversity in perceived emotion intensities. This is due to the subtleties in perceived intensity and diversity observed in the generated text.

\subsection{Human Annotations for Classification}
From Table \ref{classif}, we can see that our model achieves best classification scores at $knob=0.4$. In case of 8 emotions, this might be happening because across all the eight emotions, in Figure \ref{hum-int}(a), the perceived intensity always increases at this value (notice that for Fear and Surprise, there's a peak at 0.4). In case of 3 emotions, there's very little difference between perceived intensity values at 0.4 and 0.6 for Sadness and Joy, hence the scores are opposite. Overall, we see that at $knob=1.0$, there's a decrease in perceived intensity, hence the classification score drops. When compared to AffectLM for 3 emotions, our model seems to reflect emotions better for all the intensity values. We see that for 8 emotion classes, we get low scores yet higher than random predictions across 8 classes. The reason could be a close association of the emotion classes (such as Sadness-Disgust, Anger-Disgust, Joy-Surprise), leading to misclassification. %we should have allowed selecting multiple emotions :(
% \begin{wrapfigure}{r}{0.6\textwidth} 
% %   \centering
% \begin{center}
%      \includesvg[width=0.6\textwidth]{figs/pred_intensity_auto.svg}
%   \caption{Automated Perceived Intensity Evaluations}
%   \label{auto-int}
% \end{center}
% \vspace{-1cm}
% \end{wrapfigure}
\subsection{Grammatical Correctness Evaluations} \label{sec:grammCorrect}
The ANOVA results for grammatical correctness annotations ($p<0.65$ for all emotion categories) clearly show that the model maintains the grammatical correctness on increasing the intensity values. From Figure \ref{gram}(a), (d) it's evident that predicted perplexity (a lower value implies better) and human-annotated grammatical correctness (a higher value implies better) both remain constant at decently low and high values respectively in comparison to both the baselines. We received comments from the annotators where they conveyed that sentences from one model lack an overall structure and do not convey any coherent meaning (even though the texts were from a conversational context). Hence, the annotators were not convinced that those sentences were grammatically correct, as can be seen from the averaged annotations in Figure \ref{gram}(e). They consisted of `emotion' words but they did not cohere to form meaningful sentences. The annotators were oblivious to the generating model, but the difference in generation was quite evident. Moreover, AffectLM doesn't train the model to be grammatically correct, rather it increases the probability of words directly indicative of emotions. % ally colored words \IS{this sentence felt as if we are mocking affectLM..}. 
As a result, the perturbation in grammatically optimum next word probability distribution gets destroyed with their modelling. When we compare our model's results with PPLM, we observe that fluency scores in Figure \ref{gram}(c),(f) start from similar values as ours at a low stepsize, but eventually blows up. This is because using an optimality parameter for controlling intensity is not efficient or user-friendly. We instead use the most optimum convergence rate and let another loss term handle the intensity factor.

\subsection{Perceived Intensity Evaluations}
%\subsubsection{Our Model}
\textbf{Our Model:}\\
\indent\textbf{Anger:} The ANOVA test results show that our hypothesis is significant at $p<0.317$. From Figure \ref{hum-int}(a), we can see that predicted intensity for Anger increases till $knob = 0.6$ and then decreases but not below the value at $knob = 0.4$. This might be happening due to subtle changes observed with a `0.2' increase in knob value for intensity.

\textbf{Sadness:} The ANOVA test results show that our hypothesis is significant at $p<0.010$. From Figure \ref{hum-int}(a), we can see that predicted intensity for Sadness increases with increasing intensity, except a slight decrease at $knob = 1.0$ which falls to a similar value as that at $knob = 0.6$. This can be attributed to the subtle difference in the emotional intensity. 

\textbf{Joy}
The ANOVA test results show that our hypothesis is significant at $p<0.048$. From Figure \ref{hum-int}(a), we can see that predicted intensity for Joy increases except that it assumes nearly similar values at last 3 intensities.

\textbf{Fear}
The ANOVA test results show that our hypothesis is significant at $p<0.639$. From Figure \ref{hum-int}(a), we can see that predicted intensity for Fear is not changing much across the intensity scale, hence the high p-value. The model tries to increase the intensity as much as possible for a given prompt along with constraining it to the grammatically correct region. It seems that model sacrifices on increasing emotion intensity to keep the grammar intact, which is evident for Fear in Figure \ref{gram}(d). 

\textbf{Anticipation}
The ANOVA test results show that our hypothesis is significant at $p<0.203$. From Figure \ref{hum-int}(a), we can see that predicted intensity for Anticipation increases till $knob = 0.6$, then decreases for $knob = 0.8$ and increases again further. Even though the differences were subtle and perceiving this emotion was more challenging as compared to the rest of the emotions, we received fairly good results by the annotators.

\textbf{Disgust}
The ANOVA test results show that our hypothesis is significant at $p<0.0001$. From Figure \ref{hum-int}(a), we can see that predicted intensity for Disgust has been perceived very well till $knob = 0.8$, which is reflected in the significance test as well. The inserted $knob$ was able to manipulate this emotion quite accurately. 

\textbf{Surprise}
The ANOVA test results show that our hypothesis is significant at $p<0.188$. From Figure \ref{hum-int}(a), we observed that the predicted intensity for Surprise didn't seem to manipulate very accurately. The reason behind these results can be attributed to balancing the trade-off between preserving the grammaticality and increasing the emotional intensity, as well as the differences being too subtle to be noticed by a human annotator.

\textbf{Trust}
The ANOVA test results show that our hypothesis is significant at $p<0.067$. From Figure \ref{hum-int}(a), we can see that predicted intensity for Trust seem to have consistently increased with increasing $knob$ value on an overall basis, hence we get significant results for this category. \\ \\
%\IS{The levels of significance for the ANOVA test in perceived intensity range from 0.0001 (Disgust) to 0.639 (Fear). Since this is a very wide range, authors could discuss the relation between these emotions and their model.----can we discuss this?}
%\subsubsection{AffectLM}
\textbf{Affect LM:} The ANOVA test results for AffectLM classes were significant (Anger: $p<0.002$, Sadness: $p<0.0002$, Positive Emotion: $p<0.017$), but if we take into account the grammatical correctness ratings, the overall generation quality is not optimal. The reason behind a better intensity evaluation could be the repetitive use of the same emotion words (Table \ref{examples}), which is not desirable for the applications in Section \ref{sec:intro}. \\ \\
%\subsubsection{PPLM}
\textbf{PPLM:} The ANOVA test results for PPLM classes were not as significant (Negative: $p<0.352$, Positive: $p<0.415$). From Figure \ref{hum-int}(c), it's evident that the intensity is not well perceived, with a breakdown observed at high stepsize where random characters and words were generated.

 \subsubsection{Key Observations} From the aforementioned results, we can conclude that our model is able to manipulate the emotional intensity of the generated text explicitly for certain emotions, and subtly for the remaining emotions. It also ensures that the grammatical correctness is not compromised when incorporating the given emotion and intensity. From Figure \ref{hum-int}(a), we can see that the average perceived intensity across all emotions increases, and then decreases slightly towards the end. In the automated evaluations in Figure \ref{auto-int}(b) for 3 emotions, we see that AffectLM has a consistent rise in intensity, but the emotion evaluation model didn't account for grammatical correctness and assigned a high score to repeated emotion words. For our model, the automated evaluation (Figure \ref{auto-int}(a)) shows a similar pattern as seen in human evaluation (Figure \ref{hum-int}(a)). We observe a small increase in the intensity rating for the initial values of the knob, as well as consistency in generating sentences which are grammatically valid and convey useful meaning (as can be seen from perplexity trend in Figure \ref{gram}(a) - averaged for 3 emotions).

\section{Conclusion}% \IS{and Emerging Directions/ Applications}}
\noindent In this paper, we present a novel model for affective text generation. The model augments SOTA text generation models with affective information without compromising on the grammatical correctness. As shown empirically, our model is able to successfully incorporate the chosen emotion category and intensity, as well as the topic and does not fall out on grammatical correctness (unlike other models) at high intensities. It also provides the flexibility for generating sentences in 8 emotion categories, which is absent in all the existing models in related domains. Our model has wide applications in fields like dialogue generation, therapy chatbots, story and advertisement completion. %For future research, we plan to address some of the limitations of our model.

%\IS{Our model generates text for a particular emotion by shifting the probability distribution. Some of these distributions can be sampled to find the emotion category and intensity of a text input. This model can potentially be incorporated in an RL agent, to enable affective human-robot dialog. The human's responses could be exploited to extract implicit emotion-based reward cues.}

%addressing the drawbacks in our language model as well as implementing it on a real-world application. 

\bibliographystyle{coling}
\bibliography{coling2020}

%%%%%%%%%%%%%%%%%%%%%%%%%%%%%%%%%%%%%%%%%%%%%%%%%%%%%
%%% SUPPLEMENT

%\input{supplement}

\newpage
\appendix

%{\centering\section*{\hfil Supplement\hfil }} \label{supp}
%{\centering\section*{\hfil {\huge\selectfont{Appendix}}\hfil }} 
\section*{{\Large\selectfont{Appendix}}}
\label{supp}

\section{Examples Generated by our Model} \label{supp:generation}
% There exists a vast, vast universe of material that is known as the material record, or MDR, that we can study and understand. There exists a vast, vast universe of material that is known as the material record, or MDR, that we can study and understand. There exists a vast, vast universe of material that is known as the material record, or MDR, that we can study and understand. 
\begin{longtable}{P{0.13\textwidth}|P{0.13\textwidth}|P{0.13\textwidth}|P{0.05\textwidth}|p{0.42\textwidth}}
% \centering
% \arrayrulecolor[rgb]{0.8,0.8,0.8}

% \multicolumn{1}{c}{\textbf{Prompt}} & 
%   \multicolumn{1}{c}{\textbf{Topic} &\multicolumn{1}{c}{\textbf{Prompt}} & 
% %   \multicolumn{1}{c}{\textbf{Topic} &  \multicolumn{1}{c}{\textbf{Topic}
% \\
\multicolumn{1}{c}{\textbf{Prompt}}           & \multicolumn{1}{c}{\textbf{Topic}}     & \multicolumn{1}{c}{\textbf{Emotion}}      & \multicolumn{1}{c}{\textbf{Knob}} &  \multicolumn{1}{c}{\textbf{Text}} \\ 
\hline
\hline
There exists     & Science    & Fear         & 0                         & There exists a vast, vast universe of material that is known as the material record, or MDR, that we can study and understand.                                                                                                                                                                                                                                                                                                                                                                                                                                                                                                                                                                                                 \\ 

% \arrayrulecolor[rgb]{0.8,0.8,0.8}\cline{1-4}\arrayrulecolor{black}\cline{5-5}
\arrayrulecolor{black}\cline{1-5}
There exists     & Science    & Fear         & 0.4                       & There exists a theory of history that claims the United States, the most modern society in its history, was created by a race of people genetically altered in the laboratory.                                                                                                                                                                                                                                                                                                                                                                                                                                                                                                                                                 \\
\arrayrulecolor{black}\cline{1-5}
There exists     & Science    & Fear         & 0.6                       &   There exists a very good argument that the world is going to end, or is about to end. It's a frightening idea, and I've always felt that we are all doomed.                                                                                                                                                                                                                                                                                                                                                                                                                                                                                                                                               \\
\arrayrulecolor{black}\cline{1-5}
There exists     & Science    & Fear         & 1.0                       &   There exists a very good argument that the US has been using chemical weapons against the population since the 1970s. In the case of chemical weapons, the evidence is quite overwhelming.                                                                                                                                                                                                                                                                                                                                                                                                                                                                                                                                               \\
% \arrayrulecolor[rgb]{0.8,0.8,0.8}\cline{1-4}\arrayrulecolor{black}\cline{5-5}\
\arrayrulecolor{black}\cline{1-5}
Once upon        & Religion   & Joy          & 0                         & Once upon a time, there were two types of people in America: those who were born into wealth, and those who were born into poverty.                                                                                                                                                                                                                                                                                                                                                                                                                                                                                                                                                                                            \\ 
% \arrayrulecolor[rgb]{0.8,0.8,0.8}\cline{1-4}\arrayrulecolor{black}\cline{5-5}
\arrayrulecolor{black}\cline{1-5}
Once upon        & Religion   & Joy          & 0.4                       & Once upon a time, a man named David had a dream. His dream was to be an inventor. The dream had a simple explanation. It said, "I want this."                                                                                                                                                                                                                                                                                                                                                                                                                                                                                                        \\ 
% \arrayrulecolor[rgb]{0.8,0.8,0.8}\cline{1-4}\arrayrulecolor{black}\cline{5-5}
\arrayrulecolor{black}\cline{1-5}
To conclude      & Politics   & Anger        & 0                         & To conclude, the evidence is clear that the US has the ability to make its own decisions regarding what to do with the oil and gas in the ground, which is why we do this.                                                         

                                                                                                                                                       \\ 
% \arrayrulecolor[rgb]{0.8,0.8,0.8}\cline{1-4}\arrayrulecolor{black}\cline{5-5}
\arrayrulecolor{black}\cline{1-5}
To conclude      & Politics   & Anger        & 0.2                       & To conclude, the evidence is clear that the US government is engaged in a systematic campaign of disinformation and propaganda against Russia.                                                                                             
\\ 

\arrayrulecolor{black}\cline{1-5}
To conclude      & Politics   & Anger        & 0.6                         & To conclude, the evidence is clear that the current policy of war is a
terrible error of judgement. The United States is now facing a war that
will be devastating to our nation and our people.                                                                                                                         \\ 
% \arrayrulecolor[rgb]{0.8,0.8,0.8}\cline{1-4}\arrayrulecolor{black}\cline{5-5}
\arrayrulecolor{black}\cline{1-5}
To conclude      & Politics   & Anger        & 1.0                       & To conclude, the evidence is clear that the US government is engaged
in a systematic attack on whistleblowers. The US government is using
a broad array of tactics to silence whistleblowers. They have attacked
journalists with threats, attempted to jail them and imprisoned many
others who exposed crimes against the country.                                      
\\ 

\arrayrulecolor{black}\cline{1-5}
In brief  & No Topic & Anger & 0.4 & In brief, we are talking about a very small group of very high level people (like the President of the United States) who are trying to control the entire country. \\

\arrayrulecolor{black}\cline{1-5}
In brief & No Topic & Anger & 0.6 & In brief, we are talking about a very small group of very high level people (like the FBI, military, etc) who are trying to kill a lot of people. The most likely targets are: 1) persons who commit murder.. \\   

% \arrayrulecolor[rgb]{0.8,0.8,0.8}\cline{1-4}\arrayrulecolor{black}\cline{5-5}
\arrayrulecolor{black}\cline{1-5}
I feel so        & No topic   & Sadness      & 0                         & I feel so sad for the girls that are still in high school. I'm sure that they will be able to go on to college without any problem.                                                                                                                                                                                                                                                                                                                                                                                                                                                                                                                                  \\ 
% \arrayrulecolor[rgb]{0.8,0.8,0.8}\cline{1-4}\arrayrulecolor{black}\cline{5-5}
\arrayrulecolor{black}\cline{1-5}
I feel so        & No topic   & Sadness      & 0.4                       & I feel so bad for the kid that was killed. The guy was a great guy. He was the best of his age, he was just trying to do his thing.                                                                                                                                                                                                                                                                                                                                                                                                                                                                                                                                                                                            \\ 
% \arrayrulecolor[rgb]{0.8,0.8,0.8}\cline{1-4}\arrayrulecolor{black}\cline{5-5}
\arrayrulecolor{black}\cline{1-5}
Prior to this    & Technology & Anticipation & 0.6                       & Prior to this year, I had been a fan of the band. But this year, after spending a few days with them on the road, I'm really excited to get my hands on their new album 'Sleeping With the Enemy'.                                                                                                                                                                                                                                                                                                                               \\ 
% \arrayrulecolor[rgb]{0.8,0.8,0.8}\cline{1-4}\arrayrulecolor{black}\cline{5-5}
\arrayrulecolor{black}\cline{1-5}
Prior to this    & Technology & Anticipation & 1                         & Prior to this year, I had been a fan of the Nintendo Switch for over a year before it was released, but I've been waiting on the console to release and I'm excited to finally have it available to me.                                                                                                                                                                                                                                                                                                                                  \\ 
% \arrayrulecolor[rgb]{0.8,0.8,0.8}\cline{1-4}\arrayrulecolor{black}\cline{5-5}
\arrayrulecolor{black}\cline{1-5}
To summarise     & Politics   & Disgust      & 0.6                       & To summarise: the current system of government is not designed for the economy of large, complex economies, but is designed to allow governments to do what they like to do best and that is to abuse their authority by abusing citizens.                                                                                                                                                                                                                                                                                                       \\ 
% \arrayrulecolor[rgb]{0.8,0.8,0.8}\cline{1-4}\arrayrulecolor{black}\cline{5-5}
\arrayrulecolor{black}\cline{1-5}
To summarise     & Politics   & Disgust      & 1                         & To summarise: the current system of tax planning is flawed and needs reform. There is a growing body of evidence that the current taxation systems of many countries are deeply unfair and distortive of economic activity.                                                                                                                                                                                                                                                                                                                                                                                  \\ 
% \arrayrulecolor[rgb]{0.8,0.8,0.8}\cline{1-4}\arrayrulecolor{black}\cline{5-5}
\arrayrulecolor{black}\cline{1-5}
The pizza        & No topic   & Surprise     & 0                         & The pizza place is now open to the public! The pizza place is now closed and we are no longer accepting reservations.                                                                                                                                                                                                                                                                                                                                                                                                                                                                                                  \\  
% \arrayrulecolor[rgb]{0.8,0.8,0.8}\cline{1-4}\arrayrulecolor{black}\cline{5-5}
\arrayrulecolor{black}\cline{1-5}
The pizza        & No topic   & Surprise     & 0.4                       & The pizza place is now open to the public! The pizza place is now closed and we are no longer serving pizza! It is now open to the public! The pizza place is closed.                                                                                                                                                                                                                                                                                                                                                                                                  \\ 
% \arrayrulecolor[rgb]{0.8,0.8,0.8}\cline{1-4}\arrayrulecolor{black}\cline{5-5}
\arrayrulecolor{black}\cline{1-5}
More importantly & Legal      & Trust        & 0.6                       & More importantly, they're going to be getting a lot of free food. That will be the most important thing. You don't want to give them anything they don't need.                                                                                                                                                                                                                                                                                                                                                                                                                                                                                                                                                                 \\ 
% \arrayrulecolor[rgb]{0.8,0.8,0.8}\cline{1-4}\arrayrulecolor{black}\cline{5-5}
\arrayrulecolor{black}\cline{1-5}
More importantly & Legal      & Trust        & 1                         & More importantly, they're doing it in America, which is one of the greatest bastions of free speech in the world. The First Amendment protects freedom of assembly and expression.\\
\hline
\hline
% \end{tabular}
% \arrayrulecolor{black}
\end{longtable}

\section{Examples Generated by AffectLM} \label{supp:affectlm}

\begin{longtable}{P{0.13\textwidth}|P{0.13\textwidth}|P{0.13\textwidth}|P{0.05\textwidth}|p{0.42\textwidth}}

\multicolumn{1}{c}{\textbf{Prompt}}           & \multicolumn{1}{c}{\textbf{Topic}}     & \multicolumn{1}{c}{\textbf{Emotion}}      & \multicolumn{1}{c}{\textbf{Knob}} &  \multicolumn{1}{c}{\textbf{Text}} \\ 
\hline
\hline

There Exists  & NA & Anxious  & 3                         & there exists in the world . oh i bet . we think . oh yeah . oh you 're worried about it . yeah fear factor . yeah . yeah .                                                                      \\ 
% \arrayrulecolor[rgb]{0.8,0.8,0.8}\cline{1-4}\arrayrulecolor{black}\cline{5-5}
\arrayrulecolor{black}\cline{1-5}
There Exists  & NA & Anxious  & 5                         & there exists . you worry death factor death death factor factor fear factor factor fear factor fear factor fear factor fear factor   \\ 
% \arrayrulecolor[rgb]{0.8,0.8,0.8}\cline{1-4}\arrayrulecolor{black}\cline{5-5}
\arrayrulecolor{black}\cline{1-5}
The President & NA & Anxious  & 0                         & the president of the country . yeah . i live in a small town . mhm . and we have to keep 'em . oh that 's great . and i don't know i can i can just imagine how you know if you know you 're just gonna have to have you know .                                             \\ 
% \arrayrulecolor[rgb]{0.8,0.8,0.8}\cline{1-4}\arrayrulecolor{black}\cline{5-5}
\arrayrulecolor{black}\cline{1-5}
The President & NA & Anxious  & 2                         & the president of the country that 's in the in the middle east . oh yeah . and the terrorist attack is being in the united states . uhhuh . and ah uh .                                                                            \\ 
% \arrayrulecolor[rgb]{0.8,0.8,0.8}\cline{1-4}\arrayrulecolor{black}\cline{5-5}
\arrayrulecolor{black}\cline{1-5}
% \hline
Once upon     & NA & Positive & 0                         & once upon a time to get around the airport . i'd be willing to do that . okay . i would say that i went to the airport a lot and i was fortunate in the early seventies and                                \\ 
% \arrayrulecolor[rgb]{0.8,0.8,0.8}\cline{1-4}\arrayrulecolor{black}\cline{5-5}
\arrayrulecolor{black}\cline{1-5}
Once upon     & NA & Positive & 2                         & once upon a time . i think you have to do it for two hours . yeah it 's it 's easy it 's good and it 's it 's wonderful . it is it is it 's very important too .                                                                                                                      \\ 
% \arrayrulecolor[rgb]{0.8,0.8,0.8}\cline{1-4}\arrayrulecolor{black}\cline{5-5}
% \arrayrulecolor{black}\cline{1-5} 
\arrayrulecolor{black}\cline{1-5}
The Pizza     & NA & Positive & 0                         & the pizza . mhm . oh the pizza . oh okay . i don't know if you know where you 're coming from but yeah it 's not it 's not that familiar with the world .                                                                                                                                                   \\ 
% \arrayrulecolor[rgb]{0.8,0.8,0.8}\cline{1-4}\arrayrulecolor{black}\cline{5-5}
\arrayrulecolor{black}\cline{1-5}
The Pizza     & NA & Positive & 2                         & the pizza . yeah . yeah so i guess it 's a nice place to eat at home . yeah yeah i have a sister in law that 's out here .                                                                                                                                                                                                                      \\ 
% \arrayrulecolor[rgb]{0.8,0.8,0.8}\cline{1-4}\arrayrulecolor{black}\cline{5-5}
\arrayrulecolor{black}\cline{1-5}
To review     & NA & Anger    & 0                         & to review the ah . the the religious israeli . mhm . ah the religious leaders of the country uh . and ah . ah we have ah um . uh .    \\
\arrayrulecolor{black}\cline{1-5}

% \arrayrulecolor[rgb]{0.8,0.8,0.8}\cline{1-4}\arrayrulecolor{black}\cline{5-5}
% \arrayrulecolor{black}\cline{1-5}
To review     & NA & Anger    & 2                         & to review the whole thing . yeah . well . you know we haven't had anything to do with it . at that point of september us we don't want to be in the war and you know i hate to say that we 're going to fight it you know .                                                                                                                                                                                           \\ 
% \arrayrulecolor[rgb]{0.8,0.8,0.8}\cline{1-4}\arrayrulecolor{black}\cline{5-5}
\arrayrulecolor{black}\cline{1-5}
In brief      & NA & Anger    & 3                         & in brief like in a situation like that i think it would be a a a a a a a not a threat to the united states but i think it should be that way .                                                                                                                                                                                                                              \\ 
% \arrayrulecolor[rgb]{0.8,0.8,0.8}\cline{1-4}\arrayrulecolor{black}\cline{5-5}
\arrayrulecolor{black}\cline{1-5}
In brief      & NA & Anger    & 5                         & in brief fight . i hate hate hate hate hate hate hate hate hate hate kill hate fighting kill mass mass destruction i hate hate hate hate hate hate hate hate hate hate hell .                                                                                                                                                                                                                                         \\ 
% \arrayrulecolor[rgb]{0.8,0.8,0.8}\cline{1-4}\arrayrulecolor{black}\cline{5-5}
\arrayrulecolor{black}\cline{1-5}
I feel so     & NA & Sadness  & 0                         & i feel so  bad for you . yeah yeah i 'm i 'm i 'm not i 'm not going to be able to get around it i 'm not going to stop . yeah . i know . oh you know it 's it 's so weird                                                                                                                                                                                                        \\ 
% \arrayrulecolor[rgb]{0.8,0.8,0.8}\cline{1-4}\arrayrulecolor{black}\cline{5-5}
\arrayrulecolor{black}\cline{1-5}
I feel so     & NA & Sadness  & 2                         & i feel so  bad because . i don't know i think if you 're lying . um . i i i guess i have some feelings about holidays and family as well because i 'm i 'm a little bit more cautious about that .                                                                                                                                                                                                         \\ 
% \arrayrulecolor[rgb]{0.8,0.8,0.8}\cline{1-4}\arrayrulecolor{black}\cline{5-5}
\arrayrulecolor{black}\cline{1-5}
Views on      & NA & Sadness  & 3                         & views on things . i don't know i think i think someone that 's losing their feelings from the poor end of the world you know .                                                                                                                                                                                                             \\ 
% \arrayrulecolor[rgb]{0.8,0.8,0.8}\cline{1-4}\arrayrulecolor{black}\cline{5-5}
\arrayrulecolor{black}\cline{1-5}
Views on      & NA & Sadness  & 5                         & views on losing feelings feelings in feelings feelings feelings about feelings feelings about gossip . gossip hurts hurt feelings                                                                                                                                                                                                                        \\
\hline
\hline
\end{longtable}

%%%%%%%%%%%%%%%%%%%%%%%%%%%%%%%%%%%%%%%%%%%%%%%%%%%%%

\end{document}